\documentclass[11pt]{article}
\usepackage [final]{acl}
\usepackage{times}
\usepackage{latexsym}
\usepackage[T1]{fontenc}
\usepackage[utf8]{inputenc}
\usepackage{microtype} 
\usepackage{enumitem}
\usepackage{graphicx}
\usepackage{tabularx}
\usepackage{float}
\usepackage{booktabs}
\usepackage{xcolor}
\definecolor{aclorange}{RGB}{255, 179, 102}
\usepackage{hyperref}
\usepackage{multirow} 
\usepackage{amssymb}
\definecolor{darkblue}{rgb}{0, 0, 0.5} 
\hypersetup{
    colorlinks=true,
    citecolor=darkblue,
    linkcolor=darkblue,
    urlcolor=darkblue
}

\title{
  \textsc{EvoSpark}: Endogenous Interactive Agent Societies for \\
  Unified Long-Horizon Narrative Evolution
}

\author{
  Shiyu He\textsuperscript{1}\thanks{\hspace{1em}Equal contribution.}, 
  Minchi Kuang\textsuperscript{2}\footnotemark[1]\textsuperscript{,}\thanks{\hspace{1em}Corresponding author.}, 
  Mengxian Wang\textsuperscript{1}\footnotemark[1], 
  Bin Hu\textsuperscript{1}, 
  \textbf{Tingxiang Gu}\textsuperscript{1}, 
  \\
  \textsuperscript{1}School of Computer Science and Technology, Xinjiang University, Urumqi 830046, China \\
  \textsuperscript{2}Department of Precision Instrument, Tsinghua University, Beijing 100084, China \\
  \texttt{\{heshiyu, wangmengxian, hubin, gutingxiang\}@stu.xju.edu.cn} \\
  \texttt{kuangmc@mail.tsinghua.edu.cn}
}

\begin{document}
\maketitle

\begin{abstract}
 Realizing endogenous narrative evolution in LLM-based multi-agent systems is hindered by the inherent stochasticity of generative emergence. In particular, long-horizon simulations suffer from \textit{social memory stacking}, where conflicting relational states accumulate without resolution, and \textit{narrative-spatial dissonance}, where spatial logic detaches from the evolving plot. To bridge this gap, we propose \textsc{EvoSpark}, a framework specifically designed to sustain logically coherent long-horizon narratives within Endogenous Interactive Agent Societies. To ensure consistency, the Stratified Narrative Memory employs a Role Socio-Evolutionary Base as living cognition, dynamically metabolizing experiences to resolve historical conflicts. Complementarily, a Generative Mise-en-Sc\`ene mechanism enforces Role-Location-Plot alignment, synchronizing character presence with the narrative flow. Underpinning these is the Unified Narrative Operation Engine, which integrates an Emergent Character Grounding Protocol to transform stochastic \textit{sparking} into persistent characters. This engine establishes a substrate that expands a minimal premise into an open-ended, evolving story world. Experiments demonstrate that \textsc{EvoSpark} significantly outperforms baselines across diverse paradigms, enabling the sustained generation of expressive and coherent narrative experiences.
\end{abstract}
\section{Introduction}

The integration of Large Language Models (LLMs) into Multi-Agent Systems (MAS) has fundamentally reshaped the landscape of generative storytelling, enabling agents to simulate complex social interactions with unprecedented fluency~\cite{park2023generative,  piaoAgentSocietyLargescaleSimulation2025}. However, while current systems excel at generating short-term vignettes, achieving long-horizon story evolution---where a simulation evolves autonomously from a minimal seed into an unbounded, self-sustaining, and logically coherent narrative ecosystem---remains an elusive goal~\cite{xiaStoryWriterMultiagentFramework2025}.

As narratives expand in complexity, two critical systemic deficits emerge in current architectures, preventing the sustainability of long-term logic. First, systems suffer from social memory stacking, where the conventional append-only memory architecture leads to the accumulation of conflicting relational states (e.g., distinct memories of being both a friend and a foe), causing behavioral incoherence~\cite{platnickIDRAGIdentityRetrievalaugmented2025, zhongMemoryBankEnhancingLarge2023}. Second, text-based agents face narrative-spatial dissonance. Lacking a mechanism to synchronize narrative progression with spatial states, agents often generate ungrounded interactions that violate the essential role-location-plot logic---such as characters appearing in disjointed locations during plot-critical transitions---thereby severing the logical link between the story and its setting~\cite{ranBOOKWORLDNovelsInteractive2025, chenHAMLETHyperadaptiveAgentbased2025}.

Beyond these functional deficits, the field is structurally constrained by a paradigm schism. Traditional interactive narratives rely on rigid script adherence, ensuring logic but sacrificing autonomy~\cite{sunDramaLlamaLLMpowered2025}. Conversely, recent LLM-based simulations often prioritize open-ended emergence~\cite{park2023generative, yangOASISOpenAgent2025}, leading to uncontrollable chaotic emergence. Although recent works imply control mechanisms~\cite{hanIBSENDirectoractorAgent2024, wangStoryVerseCoauthoringDynamic2024}, existing fragmented architectures fail to support the full spectrum of narrative control necessary for diverse simulation needs---ranging from strict hierarchical planning to open-ended evolution~\cite{sunDramaLlamaLLMpowered2025}.

To address these challenges, we propose \textsc{EvoSpark}, a unified framework that integrates narrative control, cognitive evolution, and spatial grounding to foster interactive agent societies driven by endogenous character emergence. Our core contributions are:

\begin{itemize}[leftmargin=*, itemsep=0.5em]
    \item \textbf{Unified Narrative Operation Engine (NOE):} We operationalize \textit{sparking}---stochastic LLM hallucinations---not as errors, but as drivers for creativity via the \textbf{Emergent Character Grounding Protocol (ECGP)}. Through \textbf{Ontological Promotion}, the system validates and transforms fleeting narrative-induced hallucinations into persistent, legitimate main characters, effectively turning stochastic noise into structural assets for infinite world expansion.

    \item \textbf{Role Socio-Evolutionary Base (RSB):} To resolve \textit{social memory stacking}, we introduce the RSB as a mutable, \textit{living} cognition base. Unlike static or append-only RAG approaches, the RSB employs event-driven reflection to continuously metabolize experiences---updating personality, social graphs, and goals via in-place modifications. This ensures the agent's internal state evolves in real-time, maintaining consistency with the shifting socio-dynamic landscape.

    \item \textbf{Generative Mise-en-Sc\`ene (GMS):} We mitigate narrative-spatial dissonance through GMS mechanism, which acts as a virtual stage manager to enforce strict Role-Location-Plot (RLP) alignment. It dynamically synchronizes character presence and transitions with the evolving narrative flow, ensuring spatial contexts remain logically congruent with story progression.
\end{itemize}
\section{Related Work}
\label{sec:related}

\begin{figure*}[t!]
    \centering
    \includegraphics[width=\textwidth]{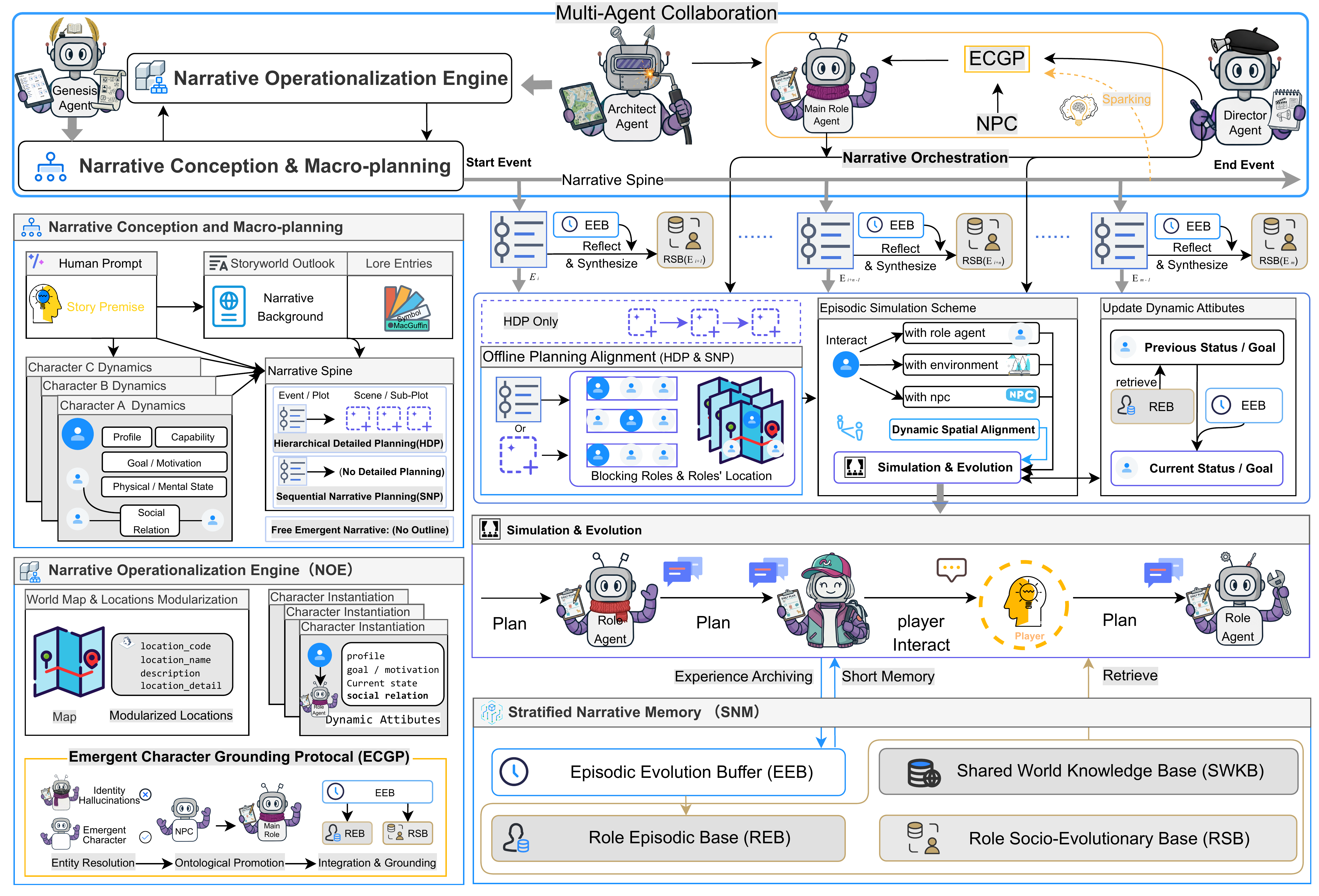}
    
    \caption{The Architecture of \textsc{EvoSpark}. 
    The framework initiates with Narrative Conception \& Macro-planning, utilizing the Unified Narrative Operation Engine for modularized storyworld and character instantiation. 
    Finally, the Simulation \& Evolution module drives the narrative loop, managing continuous interactions via the \textit{Episodic Simulation Scheme} and social memory updates based on the Stratified Narrative Memory.}
    \label{fig:framework}
\end{figure*}

\noindent \textbf{Endogenous Multi-Agent Systems} Research on MAS has transitioned from static orchestration to dynamic, endogenous evolution. Early frameworks like MetaGPT and Camel~\cite{hong2024metagpt, li2023camel} relied on fixed Standard Operating Procedures, limiting adaptability. To address this, recent architectures enable self-improvement: CoMAS and AFlow~\cite{xueCoMASCoevolvingMultiagent2025, zhang2025aflow} optimize policies via interaction rewards, while the Darwin G\"odel Machine~\cite{zhang2025darwin} allows agents to recursively modify their own code. In the realm of simulation, AgentSociety, OASIS, and Generative Agents~\cite{piaoAgentSocietyLargescaleSimulation2025, yangOASISOpenAgent2025, park2023generative} scale interactions to observe emergent norms, while BookWorld~\cite{ranBOOKWORLDNovelsInteractive2025} constructs societies directly from fictional texts. However, current endogenous systems exhibit a teleological mismatch for open-ended storytelling. First, evolutionary frameworks optimize for metric convergence (e.g., pass rates)~\cite{xueCoMASCoevolvingMultiagent2025, zhang2025aflow}, treating the stochasticity vital for narrative expansion as noise. Second, social simulators like S3 and WarAgent~\cite{gao2023s3, hua2023waragent} suffer from social memory stacking---accumulating interaction logs without a metabolic mechanism to transform transient experiences into persistent structural updates, inevitably degrading long-horizon coherence.

\vspace{0.5em}

\noindent \textbf{Socio-Evolutionary Dynamics and Memory} To support long-term narratives, research primarily targets static consistency to mitigate identity drift. Frameworks like ID-RAG~\cite{platnickIDRAGIdentityRetrievalaugmented2025} employ identity knowledge graphs to ground agent personas, while Open-Theatre~\cite{xuOpentheatreOpensourceToolkit2025a}
and MemoryBank~\cite{zhongMemoryBankEnhancingLarge2023} introduce hierarchical stores to ensure retrieval accuracy. Similarly, S3 and AgentSociety~\cite{gao2023s3, piaoAgentSocietyLargescaleSimulation2025} utilize memory modules to maintain coherent behavioral patterns across social simulations. However, these architectures typically treat memory as an accumulative log. This rigidity leads to social memory stacking, where obsolete relational states persist and conflict with new developments. While Generative Agents~\cite{park2023generative} uses reflection to synthesize observations, it focuses on preserving character states. Recent works attempt to address this: G-Memory~\cite{zhang2025gmemory} evolves hierarchies based on interaction trajectories, MemEvolve~\cite{zhang2025memevolve} meta-optimizes memory architectures, and DOME~\cite{wang2025generating} tracks temporal state changes. Despite these advances, a gap remains in enabling a socio-evolutionary metabolism that fundamentally transforms agent personality and relations over time, rather than merely accumulating context.

\vspace{0.5em}

\noindent \textbf{Generative Mise-en-Sc\`ene} Current text-based agents often suffer from Narrative-Spatial Dissonance, where generated narratives detach from coherent environmental contexts. While Generative Agents~\cite{park2023generative} grounds behavior in sandboxes and AgentSociety~\cite{piaoAgentSocietyLargescaleSimulation2025} models urban mobility via AOIs, these environments often function as passive containers. BookWorld~\cite{ranBOOKWORLDNovelsInteractive2025} advances this by introducing discrete geospatial tracking and travel constraints. Similarly, NarrativeGenie and HAMLET~\cite{kumaranNarrativeGenieGeneratingNarrative2024, chenHAMLETHyperadaptiveAgentbased2025} dynamically position props and adjudicate physical interactions to match narrative beats, while Open-Theatre and HoLLMwood~\cite{xuOpentheatreOpensourceToolkit2025a, chenHoLLMwoodUnleashingCreativity2024} explicitly define spectacle or scene boundaries to constrain agent positioning. However, these frameworks often lack the granularity to maintain essential alignment between the plot, characters, and their specific locations. Standard semantic metrics also remain blind to logical misalignments, leading to interactions that violate spatial consistency~\cite{kumaranNarrativeGenieGeneratingNarrative2024}.

\section{\textsc{EvoSpark}}
\label{sec:EvoSpark}

In this section, we elaborate on the design of \textsc{EvoSpark}. The primary motive of \textsc{EvoSpark} is to bridge the gap between static narrative planning and dynamic, open-ended agent interaction. Unlike traditional systems that isolate script control from agent emergence, \textsc{EvoSpark} functions as a holistic framework tailored for long-horizon consistency. It integrates Narrative Conception and Macro Planning, Simulation and Evolution, and crucially, acts through the Unified Narrative Operation Engine (NOE), which transforms static blueprints into long-horizon evolutionary story worlds via emergent  character grounding. The overall architecture of the \textsc{EvoSpark} framework is illustrated in Figure~\ref{fig:framework}.

\paragraph{Narrative Conception and Macro Planning}
Orchestrated by the Genesis Agent, the lifecycle initiates with the synthesis of foundational assets from a user-provided story premise. The agent generates the Storyworld Outlook, lore entries, and character dynamics. Conditioned on the selected control paradigm, these components are structured into a polymorphic Narrative Spine. This spine dictates the simulation's macro-flow, manifesting as a rigid event hierarchy (HDP), a linear plot sequence (SNP), or an open-ended null state for free emergence (Free EN).

\paragraph{Narrative Operationalization Engine}
Once the spine is established, the Narrative Operationalization Engine (NOE) structures the simulation environment. As detailed in Figure~\ref{fig:framework}, this layer performs World Map \& Locations Modularization and Character Instantiation, defining static location codes and dynamic agent attributes. Furthermore, it integrates the ECGP, which filters narrative hallucinations and executes Ontological Promotion, transforming valid \textit{sparking} roles into persistent entities within the system's cognitive base.

\paragraph{Iterative Simulation \& Evolution}
The execution phase operates under the Episodic Simulation Scheme. As shown in Figure~\ref{fig:framework}, the online simulation executes a continuous loop where Role Agents interact with the environment, Non-Player Characters (NPCs), and players. To ensure \textbf{spatial coherence}, this phase integrates the \textbf{Generative Mise-en-Sc\`ene (GMS)} mechanism. Acting as a ``Virtual Stage Manager,'' GMS bridges abstract narrative intent with concrete scene execution via a collaborative Plan-Correct protocol between the Genesis and Director agents. It enforces strict Role-Location-Plot (RLP) alignment through a dual-phase process:

\begin{figure}[t!]
    \centering
    \includegraphics[width=\columnwidth]{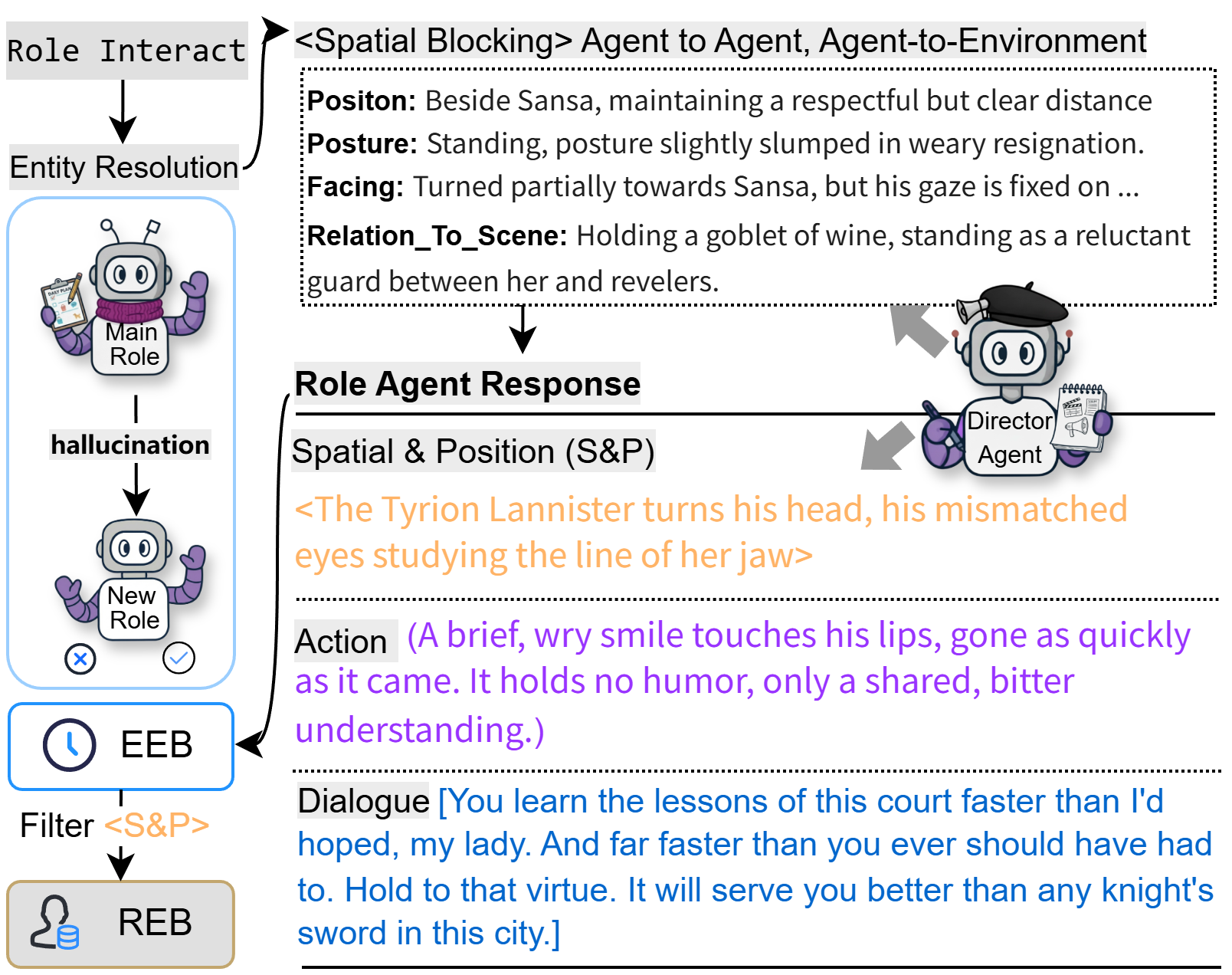}
    \caption{Dynamic Spatial Alignment. The Director Agent orchestrates narrative interactions driven by spatial context, integrating Entity Resolution and precise grounding to ensure logical consistency.}
    \label{fig:dynamic_align}
\end{figure}

\begin{itemize}[leftmargin=*, label=$\bullet$]
    \item \textbf{Offline Planning Alignment:} As detailed in Figure~\ref{fig:framework}, the Genesis Agent establishes foundational logic by aligning constraints across Role, Location, and Plot dimensions, ensuring initial assignments remain logically congruent with authorial intent.
    
    \item \textbf{Dynamic Spatial Alignment:} The Director Agent leverages spatial blocking to synchronize narrative intentions with real-time contexts. As detailed in Figure~\ref{fig:dynamic_align}, this explicitly incorporates an entity resolution step to rectify LLM-induced identity hallucinations (e.g., malformed role codes), ensuring precise character grounding.
\end{itemize}

The cycle concludes with \textit{update dynamic attributes}, where the system utilizes the RSB to consolidate the previous status and new experiences into a refreshed current status. This allows agents to assimilate narrative events into long-term cognitive and social evolution.

\subsection{Narrative Spectrum Configuration}

\textsc{EvoSpark} serves as a unified, paradigm-agnostic substrate supporting the full spectrum of control granularities. At initialization ($T_0$), the NOE is configured to one of three paradigms, dynamically aligning the narrative spine and the Director Agent's intervention policy with user intents:

\paragraph{Hierarchical Detailed Planning (HDP)}
The system operates on a rigorous event tree. The Director Agent enforces hierarchical constraints to align agent behavior with authorial intent via granular guidance, ensuring high plot fidelity without rigid hard-coding.

\paragraph{Sequential Narrative Planning (SNP)}
The system generates linear Key Nodes. Agents are motivation-driven to reach these milestones but retain improvisational freedom in path planning and interaction details between nodes.

\paragraph{Free Emergent Narrative (Free EN)}
Initialized with only foundational settings (Storyworld, Roles), the Director Agent removes plot constraints while retaining interaction guidance. Consequently, narrative trajectories are driven entirely by endogenous agent decisions and emergent conflicts.

\subsection{Multi-Agent Collaboration}

Our approach empowers the narrative to evolve not through isolated modules, but through the dynamic collaboration of four distinct types of specialized agents.

\paragraph{Genesis Agent}
As the executor of Narrative Conception, the Genesis Agent processes the human premise to generate the Narrative Spine. Crucially, it initiates the GMS Plan-Correct protocol by collaborating with the Director Agent. During the Offline Planning Alignment phase, it establishes the foundational Role-Location-Plot logic, ensuring that the initial blueprint handed off to the Architect and Director is structurally sound and logically congruent with authorial intent.

\paragraph{Architect Agent}
This agent acts as the operational core of the NOE. While it executes World Map \& Locations Modularization to instantiate the environment, its critical collaborative role lies in the ECGP. It synergizes with the Director Agent to monitor the simulation for \textit{sparking}. When the Director identifies a valid narrative hallucination, the Architect executes Ontological Promotion, transforming these fleeting mentions into legitimized entities and seamlessly integrating them into the storyworld outlook originally defined by the Genesis Agent.

\paragraph{Director Agent}
Serving as the conductor of Iterative Simulation, the Director Agent bridges the gap between static plans and dynamic execution. It orchestrates a continuous feedback loop with Role Agents to provide real-time interaction guidance based on the configured paradigm (HDP/SNP/Free EN). Simultaneously, it engages in Dynamic spatial Alignment (part of GMS), correcting potential narrative-spatial misalignments derived from the Genesis Agent's blueprint. It also acts as the primary filter for the Architect, validating whether a \textit{sparking} aligns with the current narrative flow before requesting promotion.

\paragraph{Role Agents}
These agents execute the Episodic Simulation Scheme. Driven by the Role Socio-Evolutionary Base (RSB)---their living cognition---they engage in decentralized interactions with each other and the environment. Rather than acting in isolation, they form the endogenous social graph. Their behaviors are continuously modulated by the Director's guidance, while their interaction outcomes are metabolized back into the RSB, ensuring that the collective narrative evolution is grounded in consistent, long-term character growth.

\subsection{Emergent Character Grounding Protocol (ECGP)}

The ECGP operationalizes endogenous character emergence by harnessing stochastic \textit{sparking}---narrative hallucinations of uninitialized entities. As shown in Figure~\ref{fig:framework}, the protocol captures and integrates these entities via the following pipeline:

\paragraph{Sparking via Constraint Violation}
The protocol is triggered by a generative anomaly we term \textit{sparking}. Despite strict constraints limiting selection to the existing main character list, the LLM may hallucinate an uninitialized name to bridge a narrative gap. ECGP identifies this constraint violation not as an error, but as a \textit{sparking}---a signal of latent narrative necessity.

\paragraph{Entity Resolution}
Upon detection, the Director Agent immediately intercepts the Spark to perform rigorous verification. This step acts as a filter to distinguish genuine new entities from mere aliases (e.g., nicknames or variants of existing IDs). Only when a \textit{sparking} is confirmed as structurally novel yet contextually coherent is it validated as a legitimate candidate for emergence.

\paragraph{Ontological Promotion}
This phase executes the core logic of endogenous evolution, formalizing the transition of a \textit{sparking} from a stochastic byproduct into a simulation entity. Driven by the entity's plot-criticality, this process elevates its hierarchical status, actively repurposing hallucinations from transient noise into drivers of character evolution and ecosystem expansion.

\paragraph{Integration \& Grounding}
To finalize the emergence, the Architect Agent executes structural integration. It leverages context awareness to instantiate a new profile for the promoted character within the Storyworld Outlook and initializes their cognitive state in the RSB. This process effectively anchors the generated information, transforming what was once a fleeting narrative mention into a permanent, consistent component of the world state, ready to participate in future memory formation and narrative derivation.

\subsection{Dynamic Spatial Alignment in GMS}
\label{sec:discussion}

\newcommand{\spatial}[1]{\textcolor{aclorange}{\textit{<#1>}}}

We first observe scenarios where agents exhibit spatial dissonance in the absence of the \textit{Dynamic Spatial Alignment} within GMS. As shown in Table \ref{tab:negative_example}, while the semantic plot may progress logically, physical contradictions often emerge during complex interactions, illustrating how agents can become ``lost in space'' without a persistent spatial context.

\begin{table*}[t!]
\centering
\small
\begin{tabularx}{\textwidth}{l X}
\toprule
\multicolumn{2}{l}{\textbf{Examples of Narrative-Spatial Dissonance}} \\
\midrule
\textbf{Turn} & \textbf{Interaction Detail} \\
\midrule
\textbf{T1} & \textbf{Cersei Lannister}: (Her voice is a cold, precise command as she addresses the guards without turning, standing rigidly by the hearth with her silhouette sharp against the firelight.) ``Bring me the signed contract. Then take the Stark girl to her chambers...'' \\
\addlinespace
\multicolumn{2}{c}{\textit{... (Turn 2: Cersei waits with her back to the room, demanding to see the seals and signatures) ...}} \\
\addlinespace
\textbf{T3} & \textbf{Cersei Lannister}: (A slow, dangerous \textbf{pivot brings her eyes to bear on the hesitant guard}, her hand remaining outstretched and her posture a blade of ice in the firelight.) ``My father's orders do not supersede mine in this room. The parchment. Now...'' \\
\addlinespace
\textbf{T4} & \textbf{Cersei Lannister}: (She snatches the heavy parchment from the man's trembling fingers, \textbf{though her eyes, chips of green ice, never leave Sansa's placid mask.}) ``Good. Now see to your duties, before I find a captain who knows how to obey.'' \\
\bottomrule
\end{tabularx}
\caption{A fragment exhibiting a physical contradiction in T4. The character is described as maintaining unbroken eye contact with Sansa Stark, which is logically inconsistent with her previous pivot to confront the guard in T3.}
\label{tab:negative_example}
\end{table*}

To address such discrepancies, the Dynamic-Spatial-Alignment within GMS acts as an implicit spatial perceiver to help agents ground their physical reality. Table \ref{tab:interaction_examples} illustrates interactions where character responses and dialogues are coherently anchored to their perceived physical environment through this mechanism.

\begin{table*}[t!]
\centering
\small
\begin{tabularx}{\textwidth}{l X}
\toprule
\multicolumn{2}{l}{\textbf{Examples with Dynamic Spatial Alignment}} \\
\midrule
\textbf{Turn} & \textbf{Interaction Detail} \\
\midrule
\textbf{T1} & \textbf{Tyrion Lannister}: \spatial{The Imp stands a half-step behind his wife, Sansa, his mismatched eyes scanning the preening lords and ladies.} (His hand drifts toward the small of her back, hesitating before falling away, a silent, aborted gesture of comfort. The Imp's gaze lingers for a moment on Joffrey, enthroned and sneering, before he leans infinitesimally closer to Sansa's ear, the rich velvet of his doublet brushing her arm.) ``Try not to listen to the crows, little dove. They only caw when there's carrion. Keep your eyes on me, if you can bear it. I am, at the very least, a smaller monster than most here.'' \\
\addlinespace
\textbf{T2} & \textbf{Sansa Stark}: \spatial{Standing rigidly beside Tyrion, her shoulders drawn tight.} (The Little Dove flinches almost imperceptibly at the warmth of his breath. Her gaze, which had been fixed on Joffrey with a practiced emptiness, falls to the polished floor, refusing to meet her husband's eyes.) ``As you say, my lord.'' \\
\addlinespace
\textbf{T3} & \textbf{Tyrion Lannister}: \spatial{His gaze shifts from Sansa's averted face to the celebrating court.} (The Imp gives a faint, humorless smile, a barely perceptible twitch of his lips. His attention turns outward, scanning the room with a practiced, weary eye.) ``A wise choice, my lady. The floor, at least, is honest. It makes no pretense of friendship while waiting to trip you. A virtue one can't find in many a lord here.'' \\
\addlinespace
\textbf{T4} & \textbf{Sansa Stark}: \spatial{Standing beside him, a captive bird in a gilded cage.} (The Little Dove’s fingers tighten on the silken fabric of her gown, the knuckles of her gloved hands turning white for an instant. Her gaze remains lowered, tracing the veins in the polished marble as if they were a map leading far away from this place.) ``It also keeps its counsel, my lord. A virtue rarer than honesty in this court, and far more valuable.'' \\
\bottomrule
\end{tabularx}
\caption{Examples of generated narratives with GMS. The \spatial{orange text} denotes spatial constraints, while text in parentheses indicates non-verbal actions. The Dynamic-Spatial-Alignment within GMS acts as an implicit spatial perceiver to help agents ground their physical reality.}
\label{tab:interaction_examples}
\end{table*}
\section{Stratified Narrative Memory (SNM)}
\label{sec:snm}

To support long-horizon story evolution, we introduce Stratified Narrative Memory (SNM). Unlike flat memory systems that suffer from \textit{social memory stacking}---where contradictory historical states persist---SNM adopts a layered architecture. It systematically segregates global truth, linear provenance, and evolving socio-cognitive states, ensuring agents act based on a \textit{living} cognition.

\subsection{Hierarchical Memory Architecture}
As shown in Figure~\ref{fig:framework}, SNM bridges immediate perception and persistent storage via four distinct components:

\begin{itemize}[leftmargin=*, noitemsep]
    \item \textbf{Episodic Evolution Buffer (EEB):} A \textbf{short episode memory} that caches real-time interactions ($E_i$) and sensory data, serving as a staging area before long-term metabolization.
    
    \item \textbf{Shared World Knowledge Base (SWKB):} Stores immutable \textbf{global truths} (e.g., lore, worldview), providing consistent ground truth for all agents.
    
    \item \textbf{Role Episodic Base (REB):} An immutable \textbf{experience log} for provenance tracking. It is decoupled from immediate decision-making to prevent context pollution.
    
    \item \textbf{Role Socio-Evolutionary Base (RSB):} The core \textbf{mutable storage} for the agent's current snapshot (personality, social graphs). It evolves continuously via in-place updates.
\end{itemize}

\subsection{Reflect-Synthesize-Consolidation}

To resolve conflicts between past and present states, we implement an event-driven Reflect-Synthesize-Consolidation mechanism (Figure~\ref{fig:reflect}). Instead of simply stacking memories, the system actively \textit{assimilates} data from the EEB into the persistent RSB:

\begin{figure}[t!]
    \centering
    \includegraphics[width=\columnwidth]{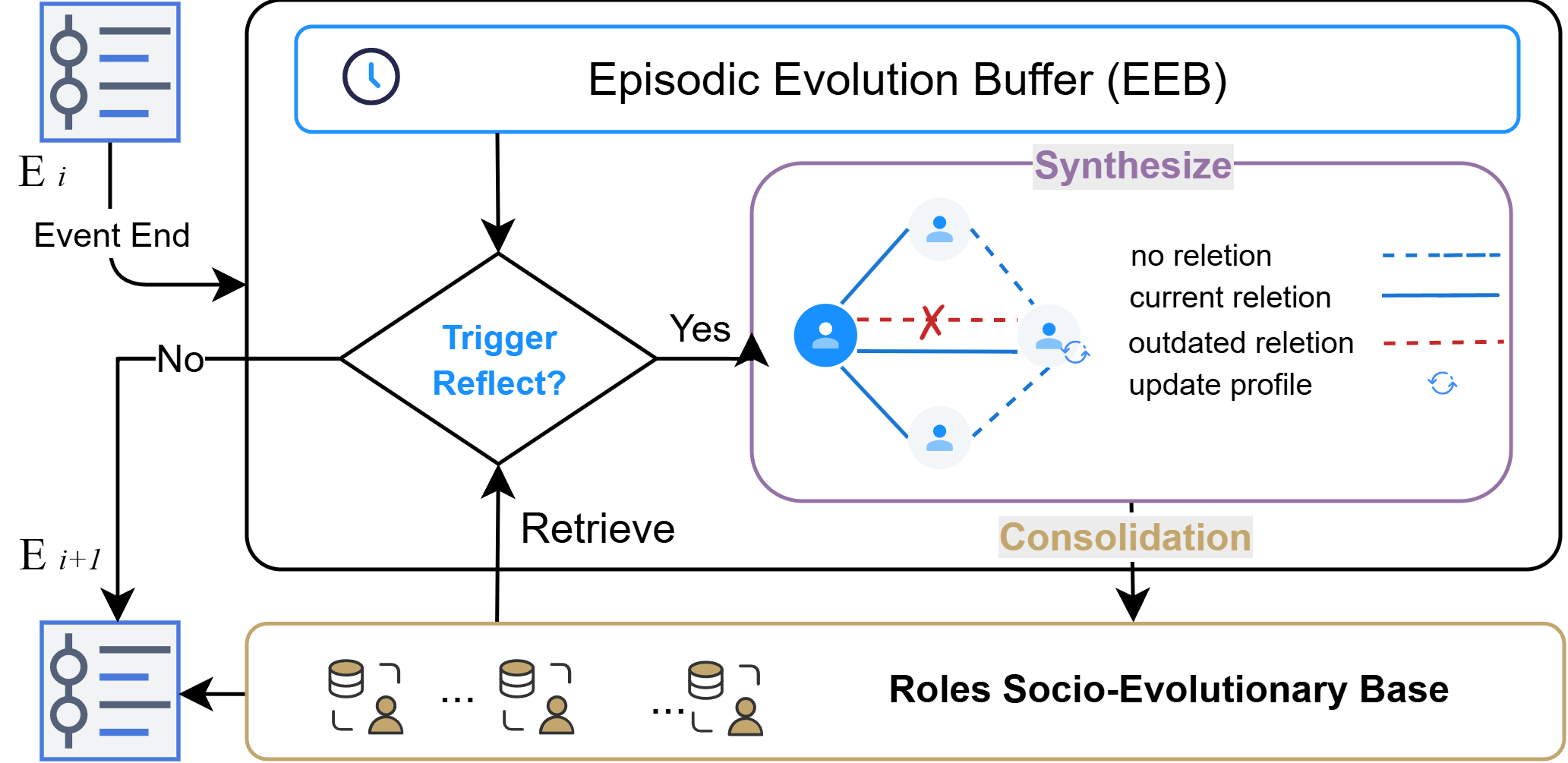}
    \caption{The event-driven Reflect-Synthesize-Consolidation mechanism.}
    \label{fig:reflect}
\end{figure}

\begin{itemize}[leftmargin=*]
    \item \textbf{Reflection Trigger:} During an event ($E_i$), raw interactions accumulate in the \textbf{EEB}. Upon event conclusion, the system executes a trigger check. If the interaction intensity exceeds a threshold, the system retrieves relevant context from the RSB; otherwise, it proceeds directly to the next event ($E_{i+1}$).

    \item \textbf{Synthesize:} This phase functions as a cognitive workspace to compute the evolutionary delta. It contrasts the emerging data cached in the EEB against established RSB states. As illustrated in Figure~\ref{fig:reflect}, the system explicitly resolves topological shifts by severing outdated relations (visualized as crossed-out edges) and synchronously updating both social connections and character profiles.

    \item \textbf{Consolidation:} The synthesized state is committed to the \textbf{RSB}. Crucially, this is an \textit{in-place update} operation. Old personality vectors and social edges are overwritten by the new synthesis, ensuring the RSB remains a consistent, conflict-free snapshot of the agent's current reality.
\end{itemize}
\section{Experiments}
\label{sec:experiments}

In this section, we validate the effectiveness of \textsc{EvoSpark} through a rigorous comparative evaluation. We assess the framework's performance across diverse narrative control paradigms, with a primary focus on the quality of story simulation and evolution.

\begin{figure*}[t!]
    \centering
    \includegraphics[width=\textwidth]{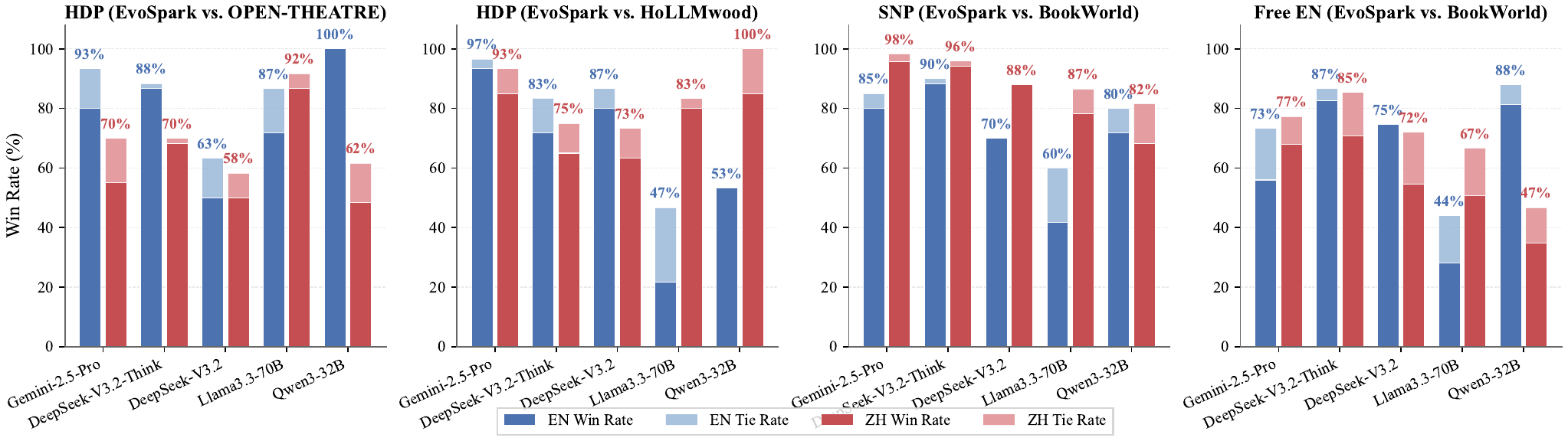}
    \caption{Comparison of win/tie rates between \textsc{EvoSpark} and baseline frameworks across different narrative modes (HDP, SNP, Free EN), languages, and LLM backbones. Detailed metric breakdowns are in Appendix \ref{sec:appendix_detailed_results}.}
    \label{fig:win_tie_rate} 
    
    \vspace{1em} 
    
    \includegraphics[width=\textwidth]{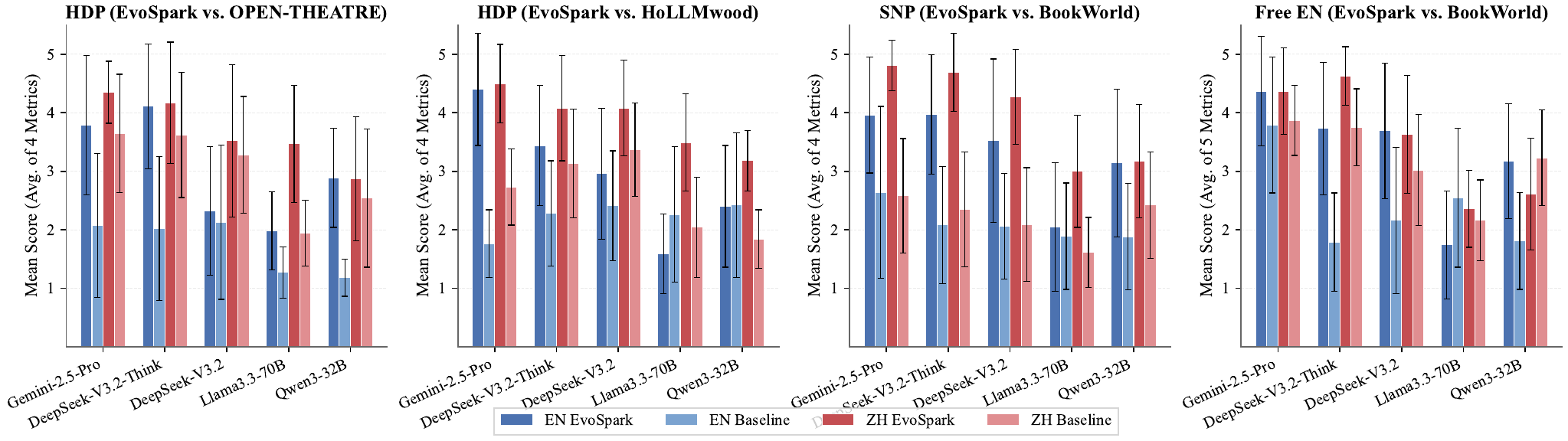}
    \caption{Comparison of overall average scores. The reported values are aggregated mean scores of underlying metrics. Detailed results are provided in Appendix~\ref{sec:appendix_detailed_results}.}
    \label{fig:mean_score}
\end{figure*}

\subsection{Experimental Setup}
\label{sec:setup}
We utilize curated scenarios covering six distinct genres (e.g., Mystery, Sci-Fi, Epic Fantasy). 
While we conduct validation across varying narrative lengths (Appendix~\ref{sec:appendix_detailed_results}), our primary evaluation prioritizes the challenging long-horizon setting within a representative subset of these genres. 
In this setting, each simulation entails a continuous sequence of 15 significant events ($\sim$45 scenes), yielding an average corpus of 200k--250k words per run to rigorously test evolutionary consistency.

\vspace{0.2em} 
\noindent\textbf{Baselines} We compare \textsc{EvoSpark} against three representative frameworks representing distinct narrative paradigms:
\begin{itemize}[leftmargin=*, nosep] 
    \item \textbf{Open-Theatre}~\cite{xuOpentheatreOpensourceToolkit2025a}: A script-driven framework utilizing a Director-Global Actor architecture, representing the centralized control paradigm.
    \item \textbf{BookWorld}~\cite{ranBOOKWORLDNovelsInteractive2025}: A virtual world simulation framework that models established environments and NPC interactions with high fidelity.
    \item \textbf{HoLLMwood}~\cite{chenHoLLMwoodUnleashingCreativity2024}: A creative writing agent framework that replicates writer-editor workflows to refine narrative quality via multi-agent collaboration.
\end{itemize}

\subsection{Evaluation Metrics}
We define a comprehensive set of metrics tailored to specific narrative paradigms.

\vspace{0.2em}
\noindent\textbf{Universal Metrics:}
\begin{itemize}[leftmargin=*, nosep]
    \item \textbf{Role Performance (RP):} Evaluates agent believability, ensuring actions and dialogues align with preset personas and evolving memories.
    \item \textbf{Immersion (Im):} Measures user engagement by assessing how effectively agents interact with the environmental context to provoke emotional resonance.
\end{itemize}

\vspace{0.2em}
\noindent\textbf{HDP \& SNP Metrics:}
\begin{itemize}[leftmargin=*, nosep]
    \item \textbf{Narrative Resonance (NR):} Evaluates structural depth. It measures thematic adherence to the blueprint and structural integrity to evoke reader empathy.
    \item \textbf{Long-Horizon Consistency (LC):} Assesses logical stability across extended segments, ensuring smooth transitions and strict adherence to the narrative spine.
\end{itemize}

\vspace{0.2em}
\noindent\textbf{Free EN Metrics:}
\begin{itemize}[leftmargin=*, nosep]
    \item \textbf{Narrative Soundness (NS):} Verifies causal feasibility, ensuring event preconditions are met and agent actions remain rational and goal-oriented.
    \item \textbf{Creativity (Cr):} Assesses content novelty. It evaluates the uniqueness of plot twists and character portrayals, penalizing generic stereotypes.
    \item \textbf{Plot Advancement (PAC):} Evaluates simulation momentum, rewarding logical conflict escalation and penalizing stagnation.
\end{itemize}

\subsection{Evaluation Methodology.}
Following BookWorld~\cite{ranBOOKWORLDNovelsInteractive2025}, we employ a pairwise \textbf{LLM-as-a-Judge} protocol with position swapping, using Gemini-2.5-Pro (English) and Deepseek-v3.2-Think (Chinese). We report Win Rates and Average Likert Scores (1--5)~\cite{likert1932technique}. The credibility of our evaluation pipeline is substantiated by its consistency with human evaluation, presented in Appendix~\ref{sec:appendix_human_eval}.

\subsection{Evaluation Results}
\label{sec:evaluation_result}
Figures~\ref{fig:win_tie_rate} and~\ref{fig:mean_score} illustrate the performance of \textsc{EvoSpark} across varying paradigms, languages, and backbones.

\paragraph{Overall Superiority on Reasoning Models.}
\textsc{EvoSpark} significantly outperforms baselines in most settings, particularly on reasoning-enhanced models (e.g., Gemini-2.5-Pro, DeepSeek-V3.2-Think). It achieves dominant win rates and margins in \textit{role performance}, \textit{narrative resonance}, and \textit{immersion}. This success is attributed to the synergy between advanced reasoning capabilities and our cognitive modules (ECGP, GMS), which demand complex instruction following to maintain consistency.

\paragraph{Complexity Gap and Stochasticity.}
Conversely, performance dips with non-reasoning models (e.g., Llama3.3-70B, Qwen3-32B) and in \textit{Free EN} mode. The former reflects an instruction-complexity gap, where \textsc{EvoSpark}'s high cognitive load (spatial/memory constraints) overwhelms weaker models. In \textit{Free EN}, volatility stems from minimal framework intervention: relaxing planning constraints increases autonomy and stochasticity, naturally causing higher variance than rigidly controlled paradigms.

\begin{figure}[t!]
    \centering
    \includegraphics[width=\columnwidth]{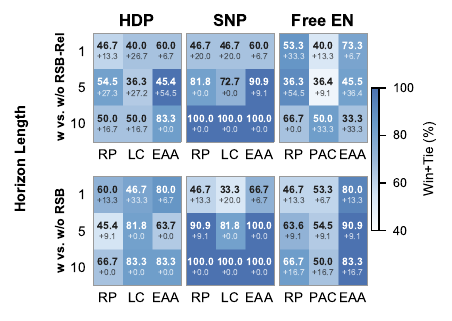}
    \caption{Long-Horizon Evolutionary Alignment Results: Win rates (bold) and tie rates of the full model vs. variants across 1, 5, and 10 events.}
    \label{fig:long_horizon_alignment}
    
    \vspace{1em}
    
    \includegraphics[width=\columnwidth]{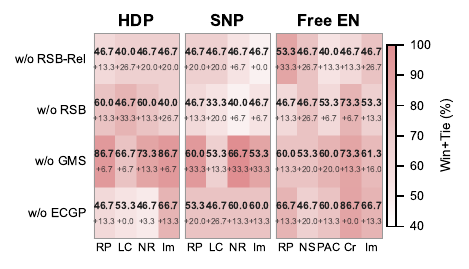}
    \caption{Ablation Study Results: Pairwise comparison heatmap between full \textsc{EvoSpark} and ablated variants. Darker red indicates higher win rates for the full model.}
    \label{fig:ablation}
\end{figure}

\subsection{Long-Horizon Evolutionary Alignment}
\label{sec:long_horizon}
Distinct from previous evaluations focused on single events, this analysis investigates system performance across continuous narrative horizons spanning 1, 5, and 10 events. 
The primary objective is to verify whether agent behaviors can dynamically shift in accordance with changes in character relationships or social identities, while maintaining long-horizon consistency. 
Central to this evaluation is \textbf{Evolutionary Action Alignment (EAA)}, a metric designed to quantify the synchronization between these evolutionary changes and agent actions.

As illustrated in Figure~\ref{fig:long_horizon_alignment}, comparisons across horizons reveal that win rates for \textbf{RP}, \textbf{LC}, and \textbf{EAA} improve significantly as the event count increases, particularly under the \textbf{SNP} and \textbf{Free EN} paradigms. 
While \textbf{HDP} also demonstrates gains, they are less pronounced than in other modes. 
We attribute this to the rigid constraints of HDP's deep plot planning, which partially attenuate the dynamic social evolution driven by the RSB.

\subsection{Ablation Study}
\label{sec:ablation}

To isolate the contributions of specific modules, we evaluated four ablated variants:
(1) \textbf{No-RSB} (excluding the Role Socio-Evolutionary Base);
(2) \textbf{No-RSB-rel} (disabling relationship evolution);
(3) \textbf{No-GMS} (omitting the Generative Mise-en-sc\`{e}ne mechanism); and
(4) \textbf{No-ECGP} (removing the Emergent Character Grounding Protocol).

Results using \texttt{gemini-2.5-pro} (Figure~\ref{fig:ablation}) align with expectations.
Removing \textbf{GMS} causes the most severe degradation across role performance, resonance, and immersion, confirming that GMS's offline and dynamic alignment is foundational for believability.

Disabling \textbf{ECGP} notably impairs immersion and creativity by restricting endogenous character emergence.
In contrast, the \textbf{No-RSB} variant exhibits a relatively marginal decline in these immediate evaluations.
We attribute this to the temporal nature of the module: the RSB is designed to mitigate cumulative memory conflicts, the effects of which become pronounced primarily over longer horizons (as detailed in Section~\ref{sec:long_horizon}) rather than in short-term comparisons.

\section{Conclusion}
We present \textsc{EvoSpark}, a framework designed to sustain logically coherent, long-horizon narratives within endogenous agent societies. 
Unlike static or script-driven approaches, our method resolves \textit{social memory stacking} via \textit{living} cognitive metabolism and mitigates \textit{narrative-spatial dissonance} through generative mise-en-sc\`{e}ne. 
Experiments demonstrate that \textsc{EvoSpark} significantly outperforms baselines in logical consistency and social fidelity across extended horizons. 
By leveraging stochastic hallucinations as structural narrative assets, our system enables the infinite expansion of open-ended story worlds. 
We hope this work paves the way for future advancements in autonomous narrative intelligence.
\section*{Limitations}
\label{sec:limitations}
Despite \textsc{EvoSpark}'s advancements in long-horizon narrative consistency, certain limitations persist. Primarily, although GMS and RSB updates are event-driven to conserve resources, the progressive accumulation of extensive narrative histories and evolving relationship graphs incurs substantial memory overhead and increased inference latency as the simulation lengthens. This currently constrains the framework's efficiency in resource-limited or strictly real-time environments. Additionally, since our current evaluation prioritizes autonomous agent-to-agent evolution to rigorously validate internal coherence, the dynamics of human-player interactivity remain less extensively quantified. The system's responsiveness to unpredictable human inputs is a critical area we intend to address in future work through dedicated optimization and user-centric evaluations.
\section*{Ethics Statement}
\label{sec:Ethics Statement}

\paragraph{Data Provenance and Synthetic Generation}
Distinct from scraped public corpora, our datasets are \textbf{synthetically generated} within the \textsc{EvoSpark} framework. We cover six narrative scenarios, with long-horizon experiments concentrating on \textit{Epic Fantasy} and \textit{Eastern Fantasy} genres. These scenarios are constructed based on specific domain constraints to simulate fictional social dynamics. Consequently, our data contains no personally identifiable information (PII) or private real-world data, eliminating risks related to privacy infringement or copyright violation of existing literary works.

\paragraph{Human Evaluation}
To validate our automatic metrics, we conducted human evaluations involving university students proficient in both English and Chinese. We strictly adhered to ethical research practices: all participants were provided with compensation well above the local minimum wage, and informed consent was obtained regarding the usage of their annotations. We maintained strict anonymity to protect annotator privacy.

\paragraph{Societal Impact and Risks}
\textsc{EvoSpark} simulates complex social dynamics and narrative evolution. We acknowledge the potential risk that the model could be misused to generate misleading content or simulate harmful social biases inherent in the training data. To mitigate this risk, we implemented safety constraints within our system prompts to filter toxic outputs. However, as with all generative agents, we emphasize that \textsc{EvoSpark} should be utilized responsibly for educational, creative, and research applications, with careful oversight required for any deployment in open-ended user interactions.

\section*{Acknowledgments}
We gratefully acknowledge the National Key Laboratory of Time and Space Information Precision Sensing at the Department of Precision Instrument, Tsinghua University, for providing the computational platform and resources that made this research possible.

\bibliography{references}
\bibliographystyle{acl_natbib}
\clearpage
\appendix

\section{Consistency with Human Evaluation}
\label{sec:appendix_human_eval}

To validate the reliability of our model-based evaluation approach, we conducted a comprehensive agreement analysis between model assessments and human assessments.

\paragraph{Setup and Methodology}
We recruited 8 human annotators to evaluate outputs from our proposed method and baselines.To ensure a robust evaluation, we randomly sampled partial data across diverse narrative genres. The evaluation was conducted using a 5-point Likert scale, where annotators and the model independently scored the generated narratives. To quantify human-model agreement, we computed Cohen's Kappa ($\kappa$) between aggregated human judgments (majority vote) and model assessments. As shown in Table~\ref{tab:human_agreement}, results indicate substantial agreement ($\kappa$ = 0.62--0.76) across all paradigms, confirming that our automated metrics align closely with human perception.

As shown in Table~\ref{tab:human_agreement}, the results indicate substantial agreement across all paradigms, confirming that our automated evaluation metrics align closely with human perception.

\begin{table}[H]
    \centering
    \small
    \renewcommand{\arraystretch}{1.2}
    \begin{tabular}{clc}
    \hline
    \textbf{Paradigm} & \textbf{Metric} & \textbf{Kappa ($\kappa$)} \\
    \hline
    \multirow{4}{*}{\textbf{HDP}} 
     & Role Performance (RP) & 0.71 \\
     & Logical Consistency (LC) & 0.69 \\
     & Narrative Resonance (NR) & 0.65 \\
     & Immersion (Im) & 0.69 \\
    \hline
    \multirow{4}{*}{\textbf{SNP}} 
     & Role Performance (RP) & 0.73 \\
     & Logical Consistency (LC) & 0.76 \\
     & Narrative Resonance (NR) & 0.68 \\
     & Immersion (Im) & 0.71 \\
    \hline
    \multirow{5}{*}{\textbf{Free EN}} 
     & Role Performance (RP) & 0.70 \\
     & Narrative Soundness (NS) & 0.68 \\
     & Creativity (Cr) & 0.62 \\
     & Immersion (Im) & 0.72 \\
     & Plot Adv. \& Conflict (PAC) & 0.73 \\
    \hline
    \end{tabular}
    \caption{Agreement between Human and Model Judges across different control paradigms.}
    \label{tab:human_agreement}
\end{table}

\section{Implementation Details and Supplementary Results}
\label{sec:appendix_detailed_results}

This appendix supplements the main experimental analysis with three key components:
(1) validation performance on standard narrative benchmarks across diverse genres, establishing the framework's versatility; and 
(2) a granular breakdown of the specific metric scores for the long-horizon experiments reported in Section~\ref{sec:evaluation_result}; and 
(3) the specific hyperparameter configurations and generation settings utilized during inference.

Specifically for the experimental setup, all language model API requests were executed using non-streaming outputs. We set the sampling \texttt{temperature} to 0.8 to strike an optimal balance between narrative creativity and logical coherence. All other decoding parameters, including \texttt{top\_p} and \texttt{top\_k}, were retained at their default values.

\subsection{Performance on Standard Benchmarks}
\label{subsec:standard_benchmark}

To verify the framework's versatility and stability across distinct narrative styles, we conducted evaluations on a standard benchmark suite covering six genres: \textit{mystery, classical drama, science fiction, modern drama, epic fantasy,} and \textit{eastern fantasy}. 

\begin{figure}[h]
    \centering
    \includegraphics[width=\columnwidth]{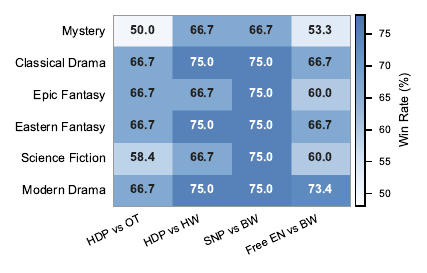}
    \caption{Cross-Domain Performance Comparison. Average win rates (\%) of \textsc{EvoSpark} against baselines across three paradigms (HDP, SNP, and Free EN) in six narrative domains. OT: OpenTheatre, HW: HoLLMwood, BW: BookWorld.}
    \label{fig:short_horizon_results}
\end{figure}

Unlike the long-horizon stress tests focused on evolutionary consistency in the main text, these simulations follow a standard episodic structure ($\sim$ 3 significant events, $\sim$6k words) to ensure broad genre coverage. 
Figure~\ref{fig:short_horizon_results} presents the comparative results. \textsc{EvoSpark} demonstrates robust performance and stylistic adaptability across all tested domains, confirming its effectiveness in handling diverse genre-specific constraints alongside complex long-term evolution.

\subsection{Granular Metric Analysis for Long-Horizon Experiments}
\label{subsec:granular_analysis}

In the main paper, we reported aggregated performance metrics to provide an overall assessment. Here, we present the comprehensive breakdown across all individual evaluation metrics.

\begin{figure*}[t!]
    \centering
    \includegraphics[width=\textwidth]{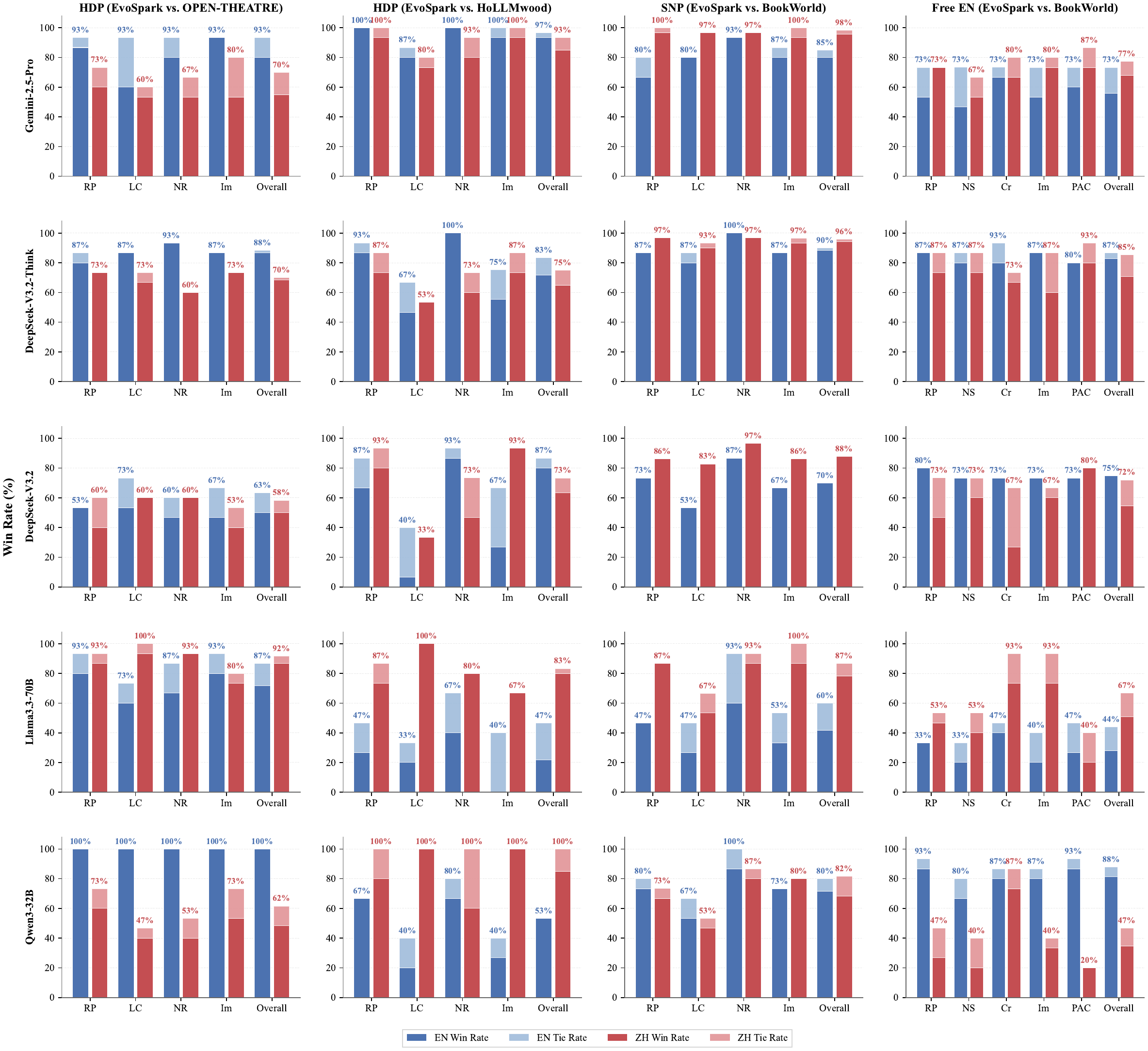} 
    \caption{Detailed Win Rates of EvoSpark vs. Baselines across all individual evaluation metrics. This breakdown covers Role Performance (RP), Immersion (Im), Narrative Resonance (NR), Long-Horizon Logical Consistency (LC), Narrative Soundness (NS), Creativity (Cr), and Plot Advancement and Conflict (PAC).}
    \label{fig:detailed_winrate}
\end{figure*}

\begin{figure*}[t!]
    \centering
    \includegraphics[width=\textwidth]{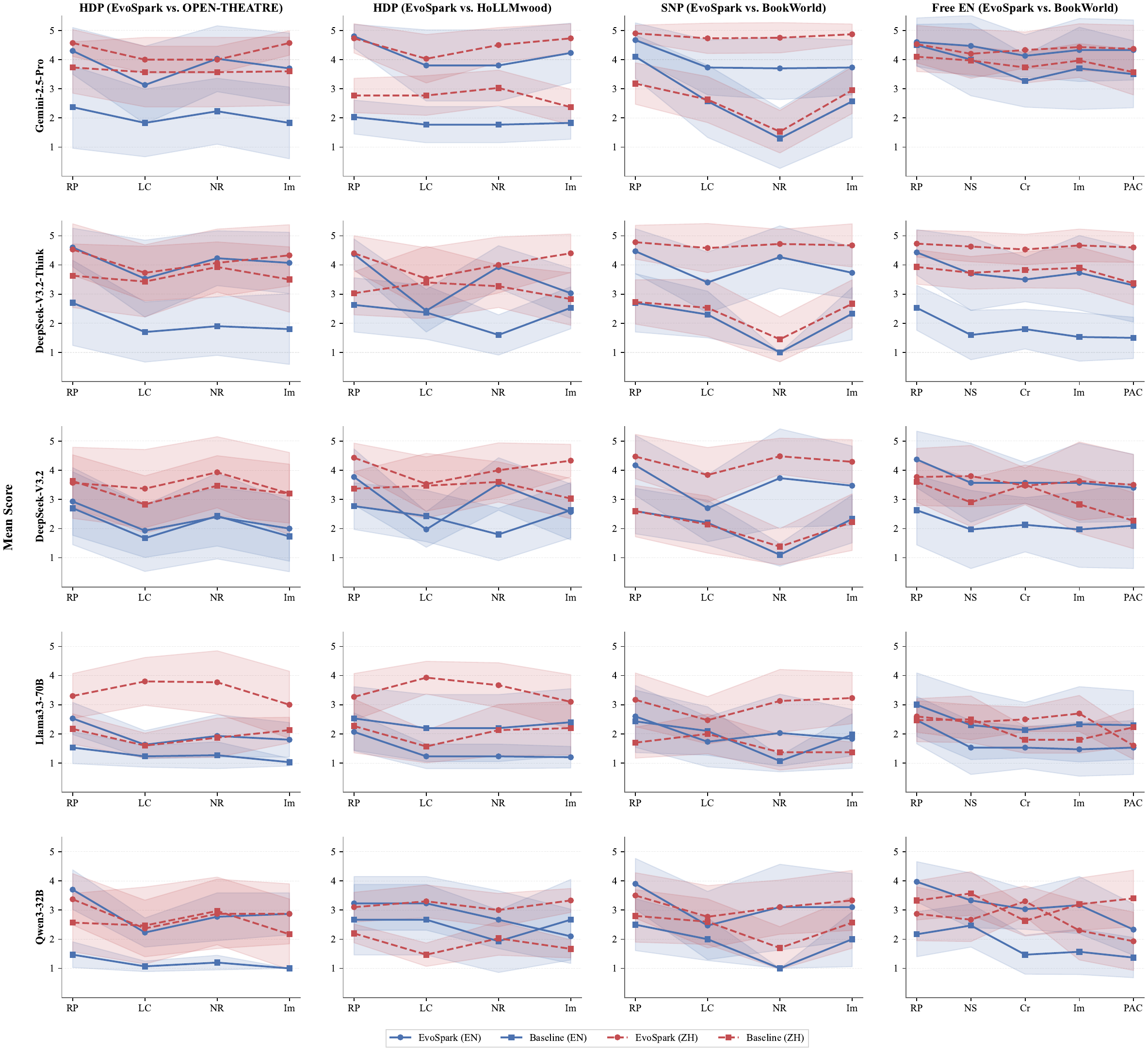} 
    \caption{Detailed Average Scores (1--5) of EvoSpark vs. Baselines across all individual evaluation metrics. The results demonstrate consistent superiority in HDP, SNP and Free EN paradigms across diverse LLM backbones.}
    \label{fig:detailed_score}
\end{figure*}

Figure~\ref{fig:detailed_winrate} illustrates the detailed pairwise Win Rates for every specific metric (RP, Im, NR, LC, NS, Cr, PAC) across different languages and backbones. 
Correspondingly, Figure~\ref{fig:detailed_score} provides the detailed Average Likert Scores (1--5). 

These granular results further substantiate that \textsc{EvoSpark}'s improvements are not limited to a single dimension but are distributed across character fidelity, spatial immersion, and narrative logic.

\subsection{Efficiency and Latency Analysis}
{Efficiency and Latency Analysis}
To investigate the computational overhead required to sustain long-horizon consistency, we evaluated the running efficiency of all systems over 100 aligned narrative steps. Specifically, this benchmark was conducted under the SNP mode utilizing the \texttt{deepseek-chat} model as the underlying reasoning engine. We define an \textit{Interaction Turn} as a complete simulation cycle where all agents perceive and act, and an \textit{LLM Call} as a single API request.

Table \ref{tab:latency_comparison} summarizes the execution overhead across all systems. While \textsc{EvoSpark} exhibits higher total duration and turn latency compared to baselines, this is primarily attributed to the multi-agent coordination required by the GMS and RSB modules to ensure narrative logic. Notably, the \textit{Avg/LLM Call} metric reveals that \textsc{EvoSpark} maintains high inference efficiency at the individual request level, suggesting that the latency trade-off is a direct result of the enhanced cognitive modeling necessary for long-term narrative evolution.

\begin{table*}[t!]
\centering
\small
\begin{tabular}{l c c c c c}
\toprule
\textbf{System} & \textbf{Total Duration} & \textbf{Avg / Turn} & \textbf{Median / Turn} & \textbf{Min / Max Turn} & \textbf{Avg / LLM Call} \\
\midrule
\textsc{EvoSpark} & 63.6 min & 38.17 s & 42.56 s & 3.99 s / 80.93 s & 3.30 s \\
OpenTheatre & 41.2 min & 24.72 s & 24.74 s & 16.51 s / 33.49 s & 7.89 s \\
BookWorld & 25.9 min & 15.53 s & 12.66 s & 2.60 s / 58.01 s & 2.15 s \\
HOLLMwood & 9.1 min & 5.46 s & 5.04 s & 2.29 s / 18.82 s & 4.77 s \\
\bottomrule
\end{tabular}
\caption{Turn Latency Comparison based on a 100-Turn Benchmark. All statistics reflect resource consumption during the evolutionary simulation. Avg / LLM Call isolates pure inference efficiency.}
\label{tab:latency_comparison}
\end{table*}

\section{Evaluation Scenario Details}
\label{sec:appendix_scenarios}

In this section, we provide the detailed specifications of the six narrative scenarios used in our experiments. Table \ref{tab:scenario_details} outlines the metadata for each domain, including genre, title, language, and source type. The specific story premises utilized for simulation initialization are detailed in the subsequent text.

\begin{table*}[t!]
    \centering
    \small 
    \renewcommand{\arraystretch}{1.3} 
    \begin{tabularx}{\textwidth}{l c c X} 
        \toprule
        \textbf{Genre (Title)} & \textbf{Lang.} & \textbf{Type} & \textbf{Story Premise} \\
        \midrule
        
        \textbf{Mystery} & ZH & Existing & 
        \textit{The Longest Day in Chang'an}. On the Lantern Festival, undercurrents surge in Chang'an as Turkic ``Wolf Guards'' infiltrate, plotting a devastating fire attack named ``Quelehuoduo.'' Li Bi, head of the Jing'an Department, recruits death-row inmate Zhang Xiaojing to save the city. Within twelve hours, this duo navigates both street chases and court politics. They uncover that the terror plot intertwines with the Crown Prince's struggle and official conspiracies aimed at the Tang Dynasty's foundation. Ultimately, they risk their lives to prevent the Lantern Tower explosion, resolving the imperial crisis. \\
        \midrule
        
        \textbf{Classical Drama} & EN & Existing & 
        \textit{Romeo and Juliet}. Amidst the feud between Montagues and Capulets in Verona, two star-crossed lovers marry in secret. A chain of tragic misunderstandings and fatal duels forces them toward a heartbreaking destiny that ultimately unites their warring families in grief. \\
        \midrule
        
        \textbf{Epic Fantasy} & EN & Existing & 
        \textit{A Song of Ice and Fire}. Following the Red Wedding, the scattered remnants of House Stark must forge new identities through trauma. They aim to dismantle the alliance of terror, reclaim Winterfell, and restore honor in a brutal saga of vengeance and rebirth. \\
        \midrule
        
        \textbf{Eastern Fantasy} & ZH & Synthesized & 
        \textit{Ten Thousand Years of Qi Refining}. Xu Yang, ostensibly a low-level cultivator, is actually a 100,000-year-old entity stronger than gods. When his sect faces destruction, he emerges to crush enemies with ``accidental'' displays of overwhelming power. Bound by a ``Source Shackle,'' he must eventually face upper-realm deities to protect his legacy. \\
        \midrule
        
        \textbf{Sci-Fi} & EN & Synthesized & 
        \textit{The War of Lost Will (2145)}. Scientist Lin Shen discovers his ``Memory Chip'' tech is corrupted by a ``Cognitive Pollution Program.'' To prevent humanity from losing free will, he teams up with a rogue AI ``Zero'' and agent Su Li to crack the code within 72 hours, battling both external enemies and his own fading memory. \\
        \midrule
        
        \textbf{Modern Drama} & ZH & Synthesized & 
        \textit{I am the Sky of Dragon City}. Huo Tian, a humiliated delivery man, awakens a ``Future Vision'' (3-second foresight). Discovering that a wealthy heir plotted his mother's illness to steal a family recipe, he uses his ability to win high-stakes bets and dismantle the heir's family empire in a story of tactical revenge. \\
        
        \bottomrule
    \end{tabularx}
    \caption{Detailed specifications of the evaluation scenarios. The dataset covers six distinct genres, comprising three existing literary works and three synthesized open-ended narratives. \textit{Lang.} denotes the simulation language (ZH: Chinese, EN: English).}
    \label{tab:scenario_details}
\end{table*}

\paragraph{Impact of Prior Knowledge}
A critical consideration in evaluating long-horizon narrative frameworks is verifying whether a system's performance relies on memorization acquired during the LLM's pre-training phase. To objectively assess genuine deductive reasoning capabilities and rule out the interference of prior knowledge, our dataset purposefully includes both \textit{Existing} scenarios (based on famous literary works, thus possessing prior knowledge) and \textit{Synthesized} scenarios (completely original narratives lacking prior knowledge).

Table \ref{tab:existing_vs_synthesized} presents a comprehensive comparative analysis of \textsc{EvoSpark}'s performance under both scenario conditions. The results detail the win rates against baselines as well as the absolute Average Likert scores achieved by \textsc{EvoSpark} itself. The varying distribution of winning conditions across different LLM backbones demonstrates that \textsc{EvoSpark}'s superiority is not strictly dependent on pre-trained memorization.

\begin{table*}[t!]
\centering
\small
\begin{tabular}{l l c c c c c}
\toprule
\multirow{2}{*}{\textbf{Model}} & \multirow{2}{*}{\textbf{Baseline}} & \multicolumn{2}{c}{\textbf{Win Rate (\%)}} & \multicolumn{2}{c}{\textbf{\textsc{EvoSpark} Avg. Likert}} & \multirow{2}{*}{\textbf{Winner}} \\
\cmidrule(lr){3-4} \cmidrule(lr){5-6}
& & \textbf{Existing} & \textbf{Synthesized} & \textbf{Existing} & \textbf{Synthesized} & \\
\midrule
\multirow{4}{*}{Gemini-2.5-Pro} 
& vs. OpenTheatre & 80.00 & 55.00 & 3.79 & 4.35 & Existing $\checkmark$ \\
& vs. HoLLMwood & 93.30 & 85.00 & 4.40 & 4.50 & Existing $\checkmark$ \\
& vs. BookWorld (SNP) & 80.00 & 95.80 & 3.96 & 4.81 & Synthesized $\checkmark$ \\
& vs. BookWorld (FEN) & 56.00 & 68.00 & 4.37 & 4.37 & Synthesized $\checkmark$ \\
\midrule
\multirow{4}{*}{DeepSeek-V3.2-Think} 
& vs. OpenTheatre & 86.70 & 68.30 & 4.11 & 4.17 & Existing $\checkmark$ \\
& vs. HoLLMwood & 71.70 & 65.00 & 3.44 & 4.08 & Existing $\checkmark$ \\
& vs. BookWorld (SNP) & 88.30 & 94.20 & 3.97 & 4.69 & Synthesized $\checkmark$ \\
& vs. BookWorld (FEN) & 82.70 & 70.70 & 3.73 & 4.63 & Existing $\checkmark$ \\
\midrule
\multirow{4}{*}{DeepSeek-V3.2} 
& vs. OpenTheatre & 50.00 & 50.00 & 2.32 & 3.52 & Tie \\
& vs. HoLLMwood & 80.00 & 63.30 & 2.96 & 4.08 & Existing $\checkmark$ \\
& vs. BookWorld (SNP) & 70.00 & 87.90 & 3.52 & 4.27 & Synthesized $\checkmark$ \\
& vs. BookWorld (FEN) & 74.70 & 54.70 & 3.69 & 3.63 & Existing $\checkmark$ \\
\midrule
\multirow{4}{*}{Llama3.3-70B} 
& vs. OpenTheatre & 71.70 & 86.70 & 1.98 & 3.47 & Synthesized $\checkmark$ \\
& vs. HoLLMwood & 21.70 & 80.00 & 1.59 & 3.49 & Synthesized $\checkmark$ \\
& vs. BookWorld (SNP) & 41.70 & 78.30 & 2.05 & 3.00 & Synthesized $\checkmark$ \\
& vs. BookWorld (FEN) & 28.00 & 50.70 & 1.74 & 2.36 & Synthesized $\checkmark$ \\
\midrule
\multirow{4}{*}{Qwen3-32B} 
& vs. OpenTheatre & 100.00 & 48.30 & 2.89 & 2.87 & Existing $\checkmark$ \\
& vs. HoLLMwood & 53.30 & 85.00 & 2.40 & 3.18 & Synthesized $\checkmark$ \\
& vs. BookWorld (SNP) & 71.70 & 68.30 & 3.14 & 3.17 & Existing $\checkmark$ \\
& vs. BookWorld (FEN) & 81.30 & 34.70 & 3.17 & 2.61 & Existing $\checkmark$ \\
\bottomrule
\end{tabular}
\caption{Comprehensive breakdown of Win Rates and \textsc{EvoSpark}'s absolute Average Likert Scores across \textit{Existing} (prior knowledge) and \textit{Synthesized} (no prior knowledge) scenarios. The results confirm that the framework's performance does not fundamentally rely on LLM pre-training memorization.}
\label{tab:existing_vs_synthesized}
\end{table*}

\section{Main Prompts Details}
\label{sec:appendix_prompts}

To facilitate reproducibility and provide transparency into the \textsc{EvoSpark} implementation, we present the core prompt templates utilized across key functional modules. Specifically, we provide the system instructions for dynamic relationship evolution (Table \ref{tab:prompt_update_relation}), long-term memory updates (Table \ref{tab:prompt_update_profile}), generative spatial blocking (Table \ref{tab:prompt_spatial}), and emergent character instantiation (Table \ref{tab:prompt_new_role}). In these templates, terms enclosed in curly braces (e.g., \texttt{\{relation\}}) denote dynamic placeholders filled by the \textsc{EvoSpark} engine during runtime based on the real-time simulation context.

\begin{table*}[t!]
    \centering
    \small
    \renewcommand{\arraystretch}{1.1}
    \begin{tabularx}{\textwidth}{|X|}
        \hline
        \textbf{UPDATE\_RELATION\_PROMPT} \\
        \hline
        You need to update your relationships with relevant characters based on the following information. \newline
        
        \textbf{\#\# Character Description} \newline
        \{role\_profile\} \newline
        
        \textbf{\#\# Character Relationships} \newline
        \{relation\} \newline
        
        \textbf{\#\# Character Current Status} \newline
        \{status\} \newline
        
        \textbf{\#\# Conversation History} \newline
        \{history\_text\} \newline
        
        \textbf{\#\# Character Relationship Update Requirements} \newline
        Please strictly follow the requirements below and return the updated relationships in JSON format: \newline
        1. \textbf{Decision Logic:} Combine the ``Character Current Status'' and ``Conversation History'' to determine whether the relationships need to be updated. Only update if there are significant interactions or changes in the dynamic. \newline
        2. \textbf{Update Strategy:} If changes are needed, please modify or supplement the original ``detail'' field content to reflect the new state. If the existing description is still accurate and sufficient, do not change it. \newline
        3. You can only modify the values of the ``relation'' and ``detail'' fields in each sub-object. \newline
        4. The value of the ``relation'' field must be a list of strings (List[str]), for example: [``new relationship1'', ``new relationship2'']. \newline
        5. The value of the ``detail'' field must be a string. \textbf{Keep it concise and summarized} (recommended 300-500 words maximum). Focus on core relationship points and recent changes; avoid lengthy historical reviews. \newline
        6. Do not change any other keys (e.g., ``ZhaoKai-en'', ``LinWanYue-en'', etc.) or the overall JSON structure. \newline
        7. Your response cannot contain any extra text or explanations besides the updated JSON. \newline
        8. You cannot delete characters, even if there is no relationship. \newline
        
        \textbf{Important: Ensure the total JSON length does not exceed the model's output limit. Prioritize JSON completeness.} \\
        \hline
    \end{tabularx}
    \caption{Prompt template for updating character relationship networks based on recent interactions.}
    \label{tab:prompt_update_relation}
\end{table*}

\begin{table*}[t!]
    \centering
    \small
    \renewcommand{\arraystretch}{1.1}
    \begin{tabularx}{\textwidth}{|X|}
        \hline
        \textbf{UPDATE\_PROFILE\_PROMPT} \\
        \hline
        You need to update the character's ``profile'' field based on the following information. \newline
        
        \textbf{\#\# Original Character Description (JSON format)} \newline
        \{role\_profile\} \newline
        
        \textbf{\#\# Character Current Status} \newline
        \{status\} \newline
        
        \textbf{\#\# Conversation History} \newline
        \{history\_text\} \newline
        
        \textbf{\#\# Character Description Update Requirements} \newline
        Please strictly follow the requirements below and \textbf{only return the updated ``profile'' field's string content}: \newline
        1. Analyze the ``profile'' field in the ``Original Character Description''. \newline
        2. Combine the ``Character Current Status'' and ``Conversation History'' to determine whether the ``profile'' field needs to be updated. \newline
        3. The ``profile'' field can only be changed when major changes related to the character occur in the story and have an impact on them. \newline
        4. If changes are needed, please modify or add to the original ``profile'' field content. \newline
        5. If no changes are needed, please \textbf{return the original ``profile'' field's string content}. \newline
        6. \textbf{Your response must be pure text string,} and can only contain the content of the ``profile'' field after updating (or without updating). \newline
        7. \textbf{Do not} include any JSON structure \newline
        8. \textbf{Do not} include any extra text or explanations (such as ``Okay, here's the updated...''). \newline
        
        For example, if the original ``profile'' is ``a student'', after updating it should become ``a student who just finished an exam'', you \textbf{can only return} ``a student who just finished an exam''. \\
        \hline
    \end{tabularx}
    \caption{Prompt template for updating character profiles (long-term memory) based on narrative progression.}
    \label{tab:prompt_update_profile}
\end{table*}

\begin{table*}[t!]
    \centering
    \small
    \renewcommand{\arraystretch}{1.1}
    \begin{tabularx}{\textwidth}{|X|}
        \hline
        \textbf{GENERATE\_SPATIAL\_POSITIONING\_PROMPT} \\
        \hline
        \textbf{\# Role Definition} \newline
        You are a professional stage director specializing in spatial blocking, skilled at arranging character positions to create dramatic tension and visual composition. \newline
        
        \textbf{\# Core Task} \newline
        Design reasonable \textbf{spatial positioning} for all participating characters (including NPCs) based on the current scene/event, character relationships, and dialogue history. \newline
        
        \textbf{\# Input Information} \newline
        \textbf{\#\# Current Scene/Event}: \{scene\_or\_event\} \newline
        \textbf{\#\# Participating Characters List}: \{roles\_list\} \newline
        \textbf{\#\# Current Location}: Location Name: \{location\_name\}; Location Description: \{location\_description\} \newline
        \textbf{\#\# Recent Dialogue History}: \{recent\_history\} \newline
        \textbf{\#\# Current Dialogue Round}: Round \{current\_round\} \newline
        
        \textbf{\# Spatial Positioning Design Principles} \newline
        \textbf{\#\# 1. Relative Position}: Describe distances (face-to-face, side-by-side...), power dynamics, emotional relationships. \newline
        \textbf{\#\# 2. Embodied Posture}: Standing, Sitting, Other postures. \newline
        \textbf{\#\# 3. Facing Direction}: Face-to-face, Back turned, Sideways, Same direction. \newline
        \textbf{\#\# 4. Scene Interaction}: Furniture interaction, Environmental interaction, Prop interaction. \newline
        
        \textbf{\# Output Format Requirements} \newline
        \textbf{Must} output a strict JSON object in the following format: \newline
        \texttt{\{\{ \newline
        ~~``spatial\_layout'': ``One-sentence description of overall spatial composition (20-40 characters)'', \newline
        ~~``positions'': \{\{ \newline
        ~~~~``Character A Name'': \{\{ \newline
        ~~~~~~``position'': ``Position in space (e.g., by window)'', \newline
        ~~~~~~``posture'': ``Body posture (e.g., standing)'', \newline
        ~~~~~~``facing'': ``Facing direction (e.g., facing Character B)'', \newline
        ~~~~~~``relation\_to\_scene'': ``Relationship to scene elements'' \newline
        ~~~~\}\}, \newline
        ~~~~... \newline
        ~~\}\} \newline
        \}\}} \newline
        
        \textbf{\# Design Considerations} \newline
        1. \textbf{Dynamic Adjustment}: Fine-tune based on dialogue development. \newline
        2. \textbf{Relationship Hints}: Use distance and facing to suggest relationships. \newline
        3. \textbf{Dramatic Tension}: Increase/reduce distance based on conflict/reconciliation. \newline
        4. \textbf{Logical Consistency}: Matching location characteristics. \newline
        5. \textbf{Include All Characters}: Ensure every participating character has clear position description. \newline
        
        --- \newline
        Now, design the spatial positioning for this round of dialogue based on the above information. \textbf{Output the JSON object directly} without any other explanations or markdown code block markers. \\
        \hline
    \end{tabularx}
    \caption{Prompt template for the Generative Mise-en-Sc\`{e}ne module, instructing the LLM to act as a stage director.}
    \label{tab:prompt_spatial}
\end{table*}

\begin{table*}[t!]
    \centering
    \small
    \renewcommand{\arraystretch}{1.1}
    \begin{tabularx}{\textwidth}{|X|}
        \hline
        \textbf{FIND\_NEW\_ROLE\_INFO\_PROMPT} \\
        \hline
        You are a skilled screenwriter. Based on the following information, generate character information for \{role\_code\}. \newline
        
        \textbf{\#\#\# Records of Previous Scenes} \newline
        \{history\_scene\_json\} \newline
        
        \textbf{\#\#\# Current Event} \newline
        \{event\} \newline
        
        \textbf{\#\#\# Information of All Other Characters} \newline
        \{role\_agents\} \newline
        
        \textbf{\#\#\# Requirements} \newline
        1. Based on the records of previous scenes, generate character information. \newline
        2. The character information should include character profile, gender, identity, and relation. \newline
        3. Return in JSON format, formatted as follows: \newline
        \texttt{\{\{ \newline
        ~~~~``profile'': ``character profile'', \newline
        ~~~~``gender'': ``character gender'', \newline
        ~~~~``identity'': ``character identity'', \newline
        ~~~~``relation'': ``character relationships'', \newline
        ~~~~``name'': ``character name'', \newline
        ~~~~``nickname'': ``character nickname'' \newline
        \}\}} \newline
        4. Forbidden to output any explanations, comments, or Markdown markers (e.g., ```json, ```python). \\
        \hline
    \end{tabularx}
    \caption{Prompt template for the Emergent Character Grounding Protocol (ECGP), used to instantiate new characters from narrative context.}
    \label{tab:prompt_new_role}
\end{table*}

\end{document}